\documentclass[10pt]{article}
\usepackage[letterpaper]{geometry}
\usepackage{hicss}
\usepackage{times}
\usepackage[none]{hyphenat}
\usepackage{url}
\usepackage{latexsym}
\usepackage{graphicx}
\usepackage{booktabs}
\graphicspath{{images/}}

\usepackage{amsmath}
\usepackage{multirow}
\usepackage[table,xcdraw]{xcolor}
\usepackage[backend=biber,style=authoryear]{biblatex}
\title{Predictive Maintenance Study  for High-Pressure Industrial Compressors: Hybrid Clustering Models}

\author{
Alessandro Costa, Emilio Mastriani, Federico Incardona, Kevin Munari, Sebastiano Spinello \\
INAF, Osservatorio Astrofisico di Catania, Via S. Sofia 78, I-95123 Catania, Italy \\
{\underline{alessandro.costa@inaf.it}, \underline{emilio.mastriani@inaf.it}, \underline{federico.incardona@inaf.it}, \underline{kevin.munari@inaf.it}, \underline{sebastiano.spinello@inaf.it}}}

\date{}

\begin{document}
\maketitle
\begin{abstract}
This study introduces a predictive maintenance strategy for high-pressure industrial compressors using sensor data and features derived from unsupervised clustering integrated into classification models. The goal is to enhance model accuracy and efficiency in detecting compressor failures. After data pre-processing, sensitive clustering parameters ($\epsilon$ and k) were tuned to identify algorithms that best capture the dataset's temporal and operational characteristics. Clustering algorithms were evaluated using quality metrics like Normalized Mutual Information (NMI) and Adjusted Rand Index (ARI), selecting those most effective at distinguishing between normal and non-normal conditions. These features enriched regression models, improving failure detection accuracy by 4.87\% on average. Although training time was reduced by 22.96\%, the decrease was not statistically significant, varying across algorithms. Cross-validation and key performance metrics confirmed the benefits of clustering-based features in predictive maintenance models.

\end{abstract}

\subsubsection*{Keywords:}

Predictive Maintenance, Industrial Compressors, Time Series Analysis, Clustering, Hybrid Clustering Modeling

\vspace{2\baselineskip}

\section{Introduction}

Predictive maintenance (PdM) represents one of the most promising asset management strategies thanks to the diffusion of Internet of Things (IoT) technologies. Leveraging a combination of data collected in real-time, the predictive maintenance approach continuously monitors the condition of equipment across all operating conditions, enabling real-time detection and analysis of potential failures. Industry studies show that the adoption of PdM can reduce failures by up to 70\%, maintenance interventions by 12\%, downtime by 50\%, and maintenance costs by up to 25\% \textcite{CapGemini2023}

The predictive maintenance market confirms this positive trend, with investments tripling from 2016 to 2021, reaching almost 4.2 billion dollars ( \textcite{marketsa2023}), and a growth projection of up to 15.6 billion dollars by 2026. This scenario is no exception for midstream oil and gas companies, where unplanned downtime can lead to productivity losses and repair costs amounting to tens of millions of dollars annually, along with the physical risks associated with equipment failure.

In light of this context, this work focuses on a high-pressure compressor equipped with 75 sensors and currently in production in a gas company. 

Recent work has highlighted the value of advanced techniques in predictive maintenance. \textcite{Hairek2023} explore the use of conditional maintenance-based fault detection and diagnosis methods, highlighting how AI techniques can optimize fault diagnosis and plant management. Furthermore, \textcite{Abassi2019} demonstrate the effectiveness of recurrent neural networks for predictive maintenance in the oil and gas industry, highlighting the benefits of advanced predictive models for equipment monitoring.

Our approach is differentiated by the use of unsupervised clustering to extract features from sensor data and improve classification models. This methodology aligns with recent innovations such as those presented by \textcite{Jambol2024}, which discuss the use of artificial intelligence for equipment management, and the work of \textcite{Chao2019} on non-destructive fault diagnostics of rotating machinery. In this context, our research seeks to further improve the performance of predictive models using selected features through clustering, thus contributing to a more effective management of industrial compressors.

Although using clustering results as additional features for further clustering models is not a very common technique, researchers are investigating this approach as a way to enhance the quality and robustness of time series analyses. For example, in \textcite{Bonacina2020} authors explore a method where clustering results, such as cluster labels, are used to enhance the understanding and analysis of time series data. This approach is particularly applied to multivariate signals in an industrial context, like monitoring a 1 MW co-generation plant. The study demonstrated that using clustering-derived features improves the effectiveness of subsequent clustering algorithms in terms of both redundancy reduction and information gain.

In  \textcite{feature-driven-time-series-clustering} scientists explore various clustering methods for time series and discuss how features derived from clustering can be employed to enhance subsequent processes, enabling the identification of more intricate patterns in time series data.

One of the advantages of using clustering results as additional features is the improved capture of temporal relationships. This means that by using cluster labels as additional features, subsequent clustering models are better able to capture temporal relationships in the data (\textcite{Zhang2021}). Additionally, the approach aids in noise reduction, as the results of initial clustering can help decrease noise in the data, enabling subsequent algorithms to focus on the most significant patterns (\textcite{Wang2020}). Moreover, by using features derived from clustering, the robustness of subsequent clustering algorithms is enhanced, particularly when dealing with complex data exhibiting high variability (\textcite{Li2022}).

The use of clustering results as additional features in time series models has the potential to significantly improve forecasting and analysis in the industrial sector. This has the potential to contribute to the continued evolution towards more intelligent and automated production processes (\textcite{Bagnall2020}).

In this paper, we propose a hybrid clustering approach to distinguish compressor operating and non-operating conditions. Such information could later help to identify failure risk events for the compressor under consideration. Starting from the raw data of the sensors, we first perform classical pre-processing tasks. Next, we study sensitive parameters such as epsilon and k, aiming to identify the clustering algorithms that best fit the data structure (both from a temporal and operational point of view). Then, coupling the pre-processed data with the temporal distinction between normal and non-normal compressor operating conditions (information provided by the site expert), we review a large set of clustering algorithms (one representative for each type) and systematically evaluate cluster quality metrics to keep the best ones as enrichment of regression models. These elements contribute to a more rigorous and comprehensive integration of clustering techniques with classification models. By first segmenting the data into clusters, this approach enhances the stability and reliability of predictive models, as it allows the classifier to exploit the inherent structure in the data identified by the clustering process..

The rest of the manuscript is structured as follows: Section 2 presents a brief review of the state-of-the-art of predictive maintenance in the petrochemical sector and a description of the compressor under study; Section 3 describes our approach to the problem; Section 4 reports some analyzes of the effectiveness of the applied strategy; finally, Section 5 proposes future developments of the work carried out.

\section{Predictive maintenance in the petrochemical sector}

Predictive maintenance represents an advanced approach to managing industrial assets in the petrochemical sector. Several studies have shown the benefits of reducing maintenance costs, improving plant reliability, and increasing overall productivity.

Research, such as that of 
\textcite{Tian2010} explored the application of condition monitoring techniques, such as vibrational analysis and thermography, to predict failures and plan targeted maintenance interventions. These sensor-based approaches have made it possible to move from reactive to more proactive maintenance.

Subsequently, the introduction of advanced machine learning and big data analytics technologies has further improved predictive capabilities. Studies such as \textcite{Lei2018} 
have shown that machine learning models, including neural networks and Random Forest, can accurately predict petrochemical plant failures.

Furthermore, the combination of process, maintenance, and fault history data has proven crucial for developing robust predictive models, as highlighted by the work of 
\textcite{Jantunen2018}.

Despite the progress made, the petrochemical sector still faces challenges related to the integration of legacy systems, the management of heterogeneous data, and the interpretability of predictive models. Recent research, such as that of \textcite{Chuka2024} 
, has explored solutions to address these issues, paving the way for further improvements in predictive maintenance in the industry.

Predictive maintenance has demonstrated significant potential in the petrochemical sector, particularly through advanced monitoring techniques and machine learning methods. For instance, vibration analysis combined with machine learning algorithms has been successfully used to detect early signs of wear in rotating equipment, allowing for timely interventions. Similarly, temperature and pressure sensors in pipelines can be analyzed using predictive models to forecast potential failures before they occur. While these techniques are already in use in some cases, the broader integration of such systems into existing operations remains challenging. Issues such as ensuring interoperability between different systems and effectively managing the vast amounts of sensor data generated require ongoing research. Future efforts should focus on refining these techniques and developing new methods to enhance predictive maintenance, ensuring they can be more seamlessly incorporated into petrochemical processes.

\subsection{The compressor in our study}

The compressor we refer to in this study,  is a robust and reliable unit, designed for high-pressure air compression applications typical of petrochemical and refining plants. Thanks to its 2-stage design, which involves compressing air in two stages for enhanced efficiency and higher pressure output, this compressor can reach delivery pressures of up to 40 bar. This makes it particularly well-suited for processes requiring high-pressure compressed air. In the first stage, air is drawn in and compressed to an intermediate pressure, then cooled before being compressed again in the second stage to reach the desired high pressure (\textcite{FluidAireDynamics}).

To ensure maximum reliability and availability, the compressor is equipped with a comprehensive set of integrated sensors, including those for monitoring vibration, temperature, pressure, and flow. These sensors cover critical points such as bearings, shafts, engine components, and compressor output.

Thanks to this rich set of sensors, the compressor is particularly well suited to the implementation of predictive maintenance strategies in petrochemical plants. The data collected in real-time can be analyzed with advanced techniques to identify any anomalies or deterioration early, allowing maintenance interventions to be planned proactively.

This is essential to guarantee the operational continuity of a critical asset such as the studied compressor, whose unexpected downtime could have serious repercussions on the entire production. By analyzing sensor data for predictive maintenance, we can optimize unit life cycle management, reduce maintenance costs, and maximize plant availability.

\section{Strategy description}

The initial raw data encompasses 8 months of compressor operation, comprising a total of 675,660 observations and 37 features. These features include a variety of sensor readings, all of which are numerical. For instance, the dataset records Suction Pressure (measured in bar), Discharge Pressure (measured in bar), and Lube Oil Pressure (measured in bar) as key pressure indicators, alongside Packing Temperature - Cylinder 1 and Packing Temperature - Cylinder 2 (both measured in degrees Celsius) for temperature monitoring. This diverse set of features allows for comprehensive monitoring of the compressor's operational status.

As shown in the diagram of (Figure 1), after the data are loaded, the analysis process involves a series of preprocessing operations, including the drop of rows with missing or invalid data.

To identify the features that should be removed, the process involves calculating the auto-correlation matrix (\cite{Brockwell2016}) and then applying the analysis of variance 
method to pinpoint the variables that exhibit the most significant discriminating power.

The data frame object in this study is accompanied by a Normal Operating Condition (NoC) file. This file identifies the periods during which the compressor was considered to be in normal operating conditions by field experts. The NoC file was used for two main purposes:
\begin{enumerate}
    \item Identifying the periods of operation and inactivity of the compressor and treating them as a sequence of sets along the time axis to quantify the effectiveness of the various clustering algorithms used (refer to Clustering and Metrics Section).
    \item Using the compressor operating state (normal/abnormal) as a target feature when training clustering models (refer to Clustering Models Section).
\end{enumerate}
Concerning our dataset, around 83.33\% of the data points are labeled as NOC, whereas approximately 16.67\% are labeled as non-NOC.
Once the features are selected, the data undergoes standardization 
, and the target variable is established based on the specific time slots (timestamps) at which the event of interest occurs. These time slots are defined in the NoC file that accompanies the dataset.

For the estimation of the hyper-parameters, three subsets of the starting dataset are defined, respectively equal to 10\%, 20\%, and 30\% of the starting data frame. The NearestNeighbors library (\textcite{scikitLearn}) is then used to obtain a plot that identifies the point of greatest curvature (epsilon neighborhood). 

The Silhouette metric 
is also calculated for the three subsets to determine the optimal clustering values.

The dataset is then divided into temporal clusters differentiated by the machine status, and the clustering algorithms are applied using the identified parameters.

Finally, ARI (Adjusted Rand Index) and NMI (Normalized Mutual Information) (\textcite{Vinh2010}) metrics are used to identify the algorithms that best represent the distribution of the dataset.
The next phase involves the implementation of clustering models (\textcite{Tan2019}), both with and without the inclusion of enriched features derived from the previous steps. Additionally, cross-validation technique 
will be used to assess the respective goodness metrics.  The time required to train the model has been estimated for both scenarios. The next subsections describe the details and results of each step.

\begin{figure}
\centering    \includegraphics[width=1\linewidth]
    {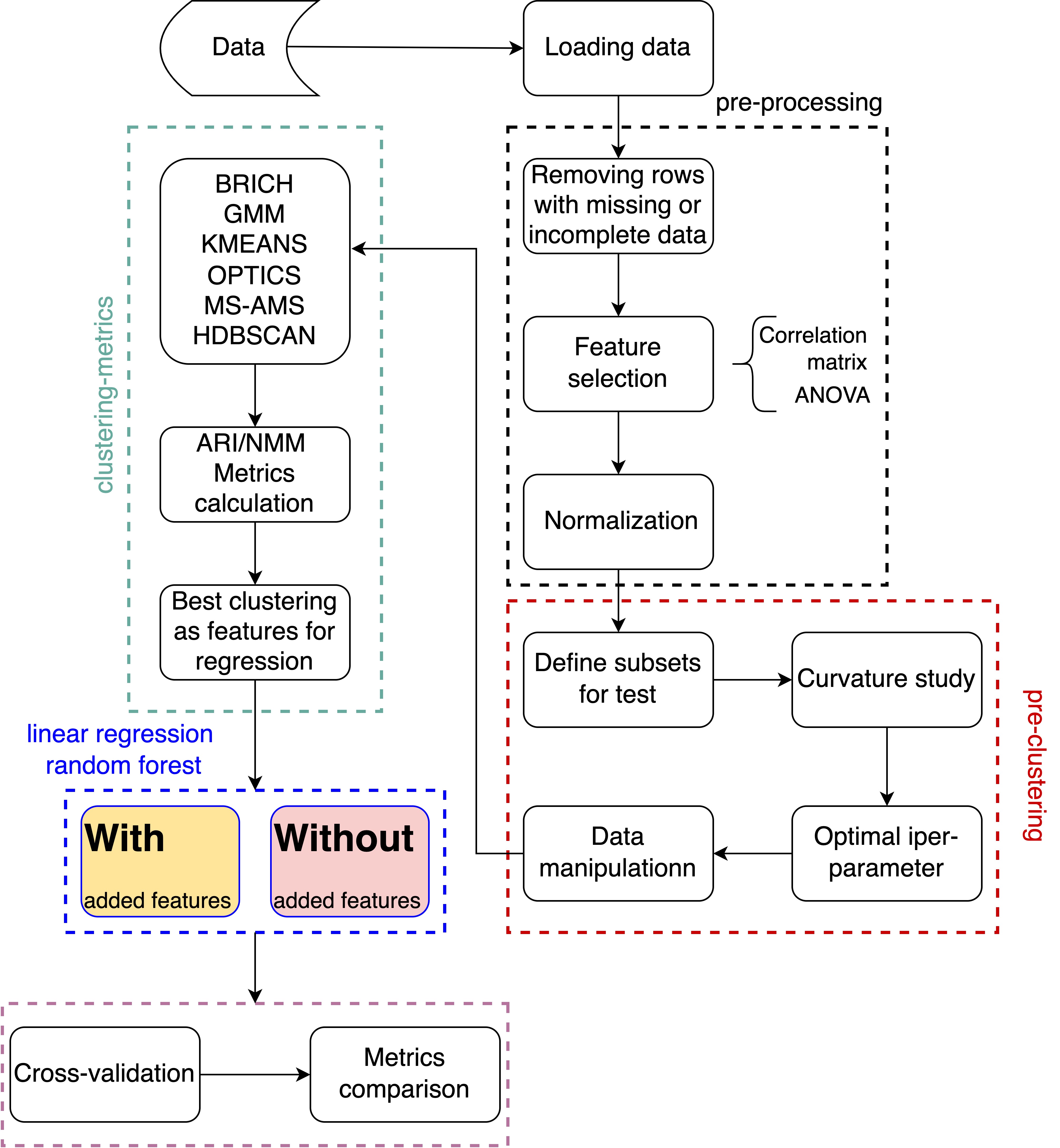}
    \caption{Flowchart of the main process. The process consists of five tasks: preprocessing the dataset, determining optimal clustering parameters, performing and evaluating clustering, applying classification algorithms, and validating with metrics.
}
\label{fig:enter-label}
    
\end{figure}
\subsection{Pre-processing}
In the realm of spatio-temporal data processing, meticulous data cleaning and preparation are essential. Initially, our focus was on identifying and eliminating rows that contained missing or invalid data, ensuring the dataset's integrity. Subsequently, we delved into calculating the auto-correlation matrix to pinpoint highly correlated features. As mention in \textcite{Kuhn2013}, any features with an absolute correlation exceeding 0.8 were systematically removed to prevent the introduction of redundancy into the model (Figure 2)

\begin{figure}
    \centering
    \vspace{-10pt} 
    \includegraphics[width=1\linewidth]
    {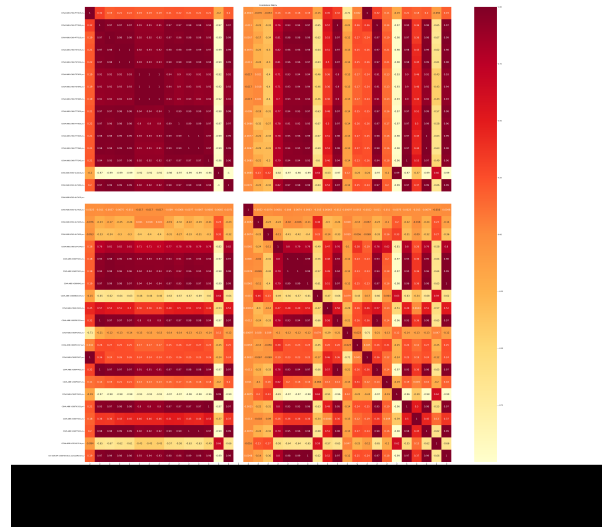}
    \vspace{-10pt} 
    \caption{Correlation matrix. The heatmap illustrates the correlation between the features in the data frame. It was generated using the corr() function in pandas.
}
    \label{fig:enter-label}
\end{figure}

Following the initial selection based on correlation analysis, we conducted an in-depth analysis of variance (ANOVA) to further evaluate the discriminant power of the remaining features (Figure 3). This process involved calculating the F-value and p-value for each feature. Only those features with a p-value below 0.05 were considered statistically significant and retained for further analysis (\textcite{James2013}). As shown in Figure 3, approximately a dozen features met these criteria, which represent the most relevant variables after filtering based on correlation and statistical significance.
Lastly, we standardized the data to ensure a consistent scale and created the target variable named "NORMAL" based on the specific time slots in which the event of interest occurred. This meticulous data preparation process is widely acknowledged as indispensable for yielding reliable and resilient results in spatio-temporal analysis.

\begin{figure}
    \centering
    \vspace{-10pt} 
    \includegraphics[width=1\linewidth]{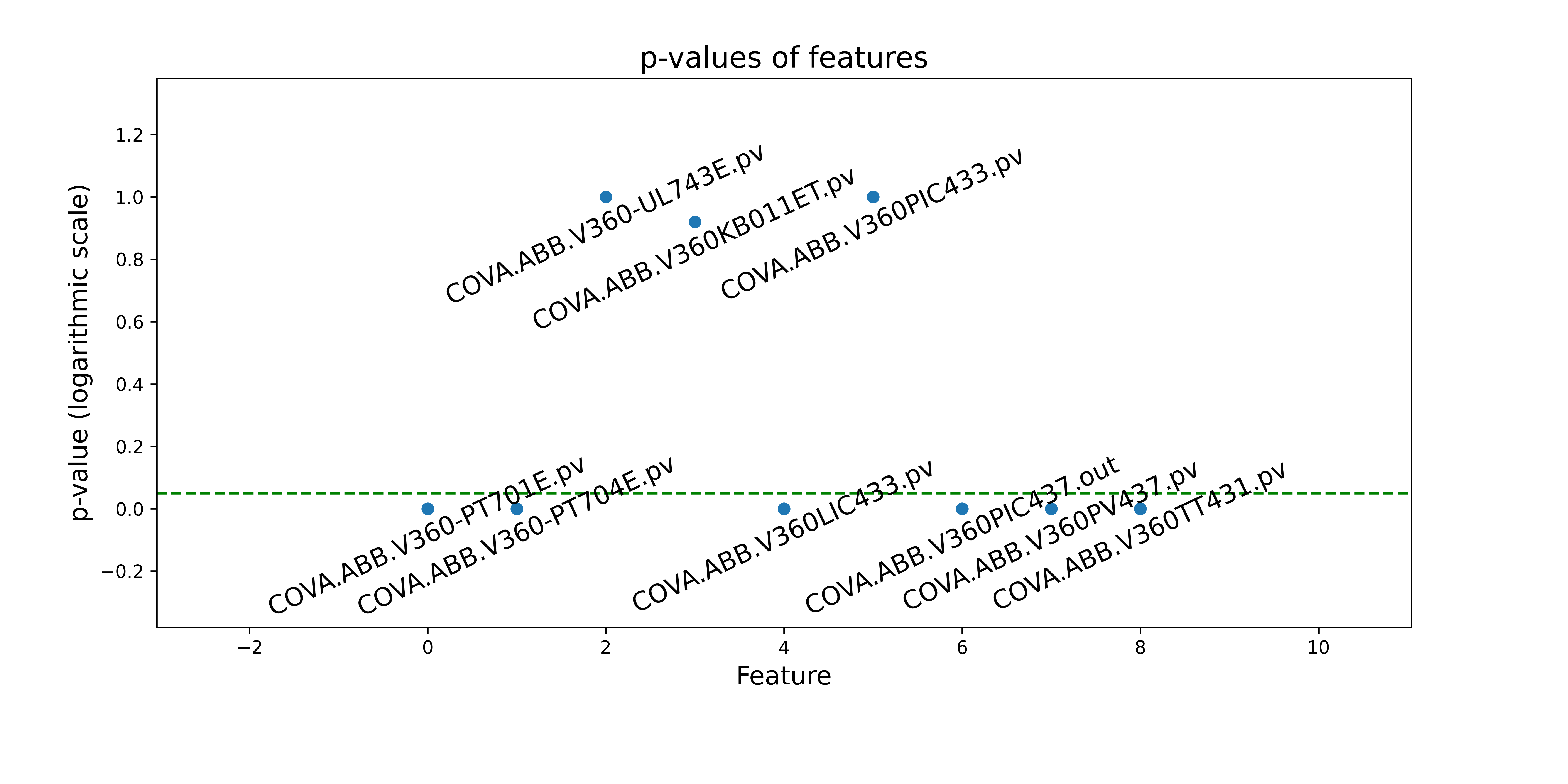}
    \vspace{-10pt} 
    \caption{P-values for features. In the scatter-plot, the features with a p-value less than 0.05 are labeled, with the threshold indicated by the green dotted line.}
    \label{fig:enter-label}
\end{figure}

\subsection{Pre-clustering}
The epsilon value is a critical parameter for density-based clustering algorithms, such as HDBSCAN (\textcite{McInnes2017}). It defines the radius within which neighboring points are considered to determine the local density. An epsilon value that is too small can lead to cluster fragmentation, while an epsilon value that is too large can cause distinct clusters to merge. Therefore, identifying the optimal epsilon value is critical for correct data segmentation.

Likewise, determining the optimal number of clusters is essential for algorithms such as K-Means. 
The wrong number of clusters can lead to misleading results, with clusters not correctly representing the structure of the data. There are several metrics, such as the Silhouette index or the elbow method 
, that can help identify the optimal number of clusters.

These critical parameters are important for K-Means and HDBSCAN and other clustering algorithms such as Gaussian Mixture Models (GMM), Adaptive Mean Shift (MS-AMS),  Ordering Points To Identify the Clustering Structure (OPTICS), and  Balanced Iterative Reducing and Clustering using Hierarchies (BIRCH). For example, GMM (\textcite{Reynolds2015}) requires specification of the number of components (clusters) to be identified, while MS-AMS (\textcite{Comaniciu2002}) requires bandwidth definition, which influences the size of the identified clusters. OPTICS 
uses the epsilon parameter to define the data clustering density, BIRCH 
uses the branching factor and threshold to determine the granularity of the clusters.

In summary, accurately identifying the epsilon value and the optimal number of clusters is critical to obtain reliable and meaningful clustering results, regardless of the algorithm used. This is a priori knowledge allowing to maximize the effectiveness of clustering algorithms in representing the true structure of the data.

To determine the optimal values of epsilon and the number of clusters, we adopted a structured three-step approach.

First, we subdivided the dataset into subsets of increasing size (10\%, 20\%, and 30\%). This allowed us to evaluate how clustering algorithms' performance varies with dataset size, and working on smaller subsets facilitated parameter exploration and optimization.

Next, we calculated the point of greatest curvature in the Nearest Neighbors graph. According to \textcite{rahmah2016determination} this technique involves analyzing the graph representing the distance of a point to its k-th nearest neighbors. The point of greatest curvature in this plot can suggest the optimal epsilon value for density-based algorithms like HDBSCAN (see Figure 4).

Finally, we identified the optimal epsilon and number of clusters (ncluster) values. After the Nearest Neighbors provided the necessary distances for calculating cohesion and separation metrics, we used the Silhouette function metric to determine the epsilon and ncluster values that maximize these metrics (see Figure 5).
\begin{figure}
    \centering
    \vspace{-10pt} 
    \includegraphics[width=1\linewidth]{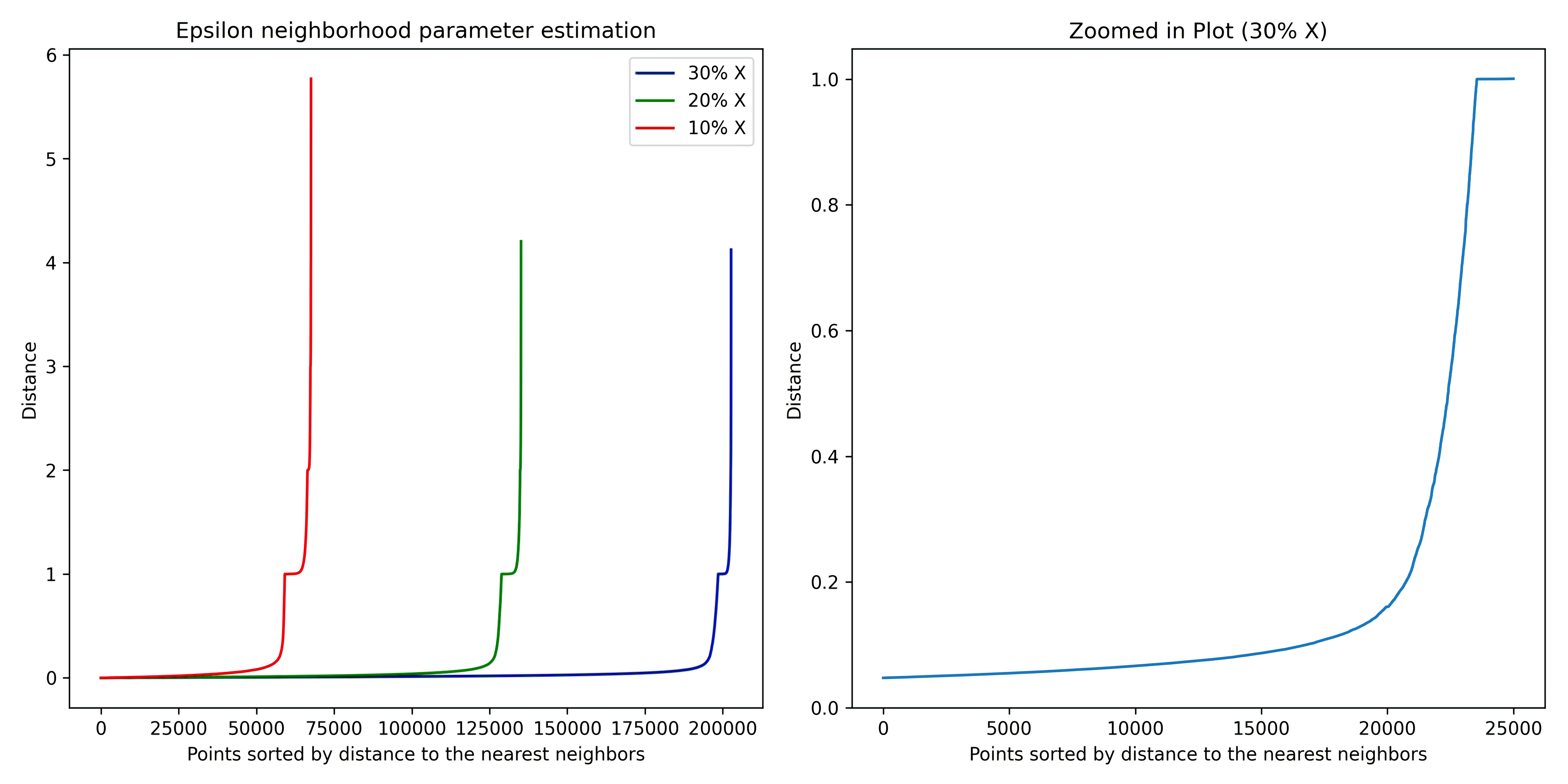}
    \vspace{-10pt} 
    
    \caption{The distance to each point's k-th nearest neighbors is averaged and plotted in ascending order. The optimal epsilon value corresponds to the point of maximum curvature on the graph.}
    \label{fig:enter-label}
\end{figure}
\begin{figure}
    \centering
    \includegraphics[width=1\linewidth]{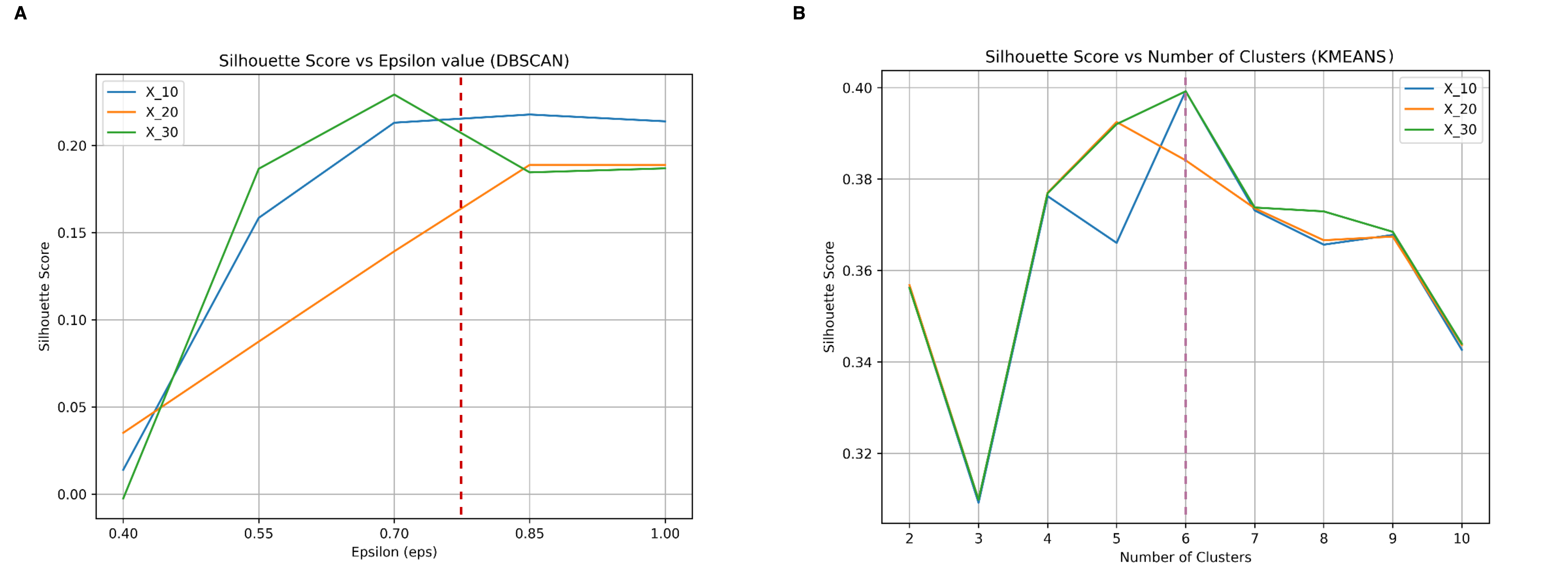}
    \caption{Silhouette evaluation: A. Silhouette Score vs Epsilon (DBSCAN) shows how clustering quality varies with different epsilon values in a density-based algorithm. B. Silhouette Score vs Number of Clusters (KMeans) illustrates how clustering quality changes with varying k values in a distance-based algorithm.}
    \label{fig:enter-label}
\end{figure}
In our study, we employed a pre-clustering strategy on the entire dataset, combining both test and training data, to achieve a comprehensive understanding of the data's global structure. Although this approach introduces a risk of data leakage, it was chosen to ensure that the clustering reflected the full dataset's characteristics, revealing patterns that might be obscured if data were partitioned beforehand. We took measures to mitigate this risk by ensuring that the final classification model was trained and validated separately, preventing any direct transfer of information from the test set. This approach provided a more accurate segmentation and valuable insights for subsequent analysis and model development.

\subsection{Clustering and Metrics}
In this task, we used several clustering algorithms configured with the previously optimized parameters (number of clusters=6, epsilon=0.8). The clustering algorithms include K-Means, HDBSCAN, OPTICS, BIRCH, Gaussian Mixture Model, and MS-AMS clustering. The results of the clustering versus time axis are displayed in Figure 6. We assigned a specific label to each period of the dataset based on the information provided by the NoC, resembling a supervised clustering process, as depicted in Figure 7.  This step allowed us to visually and analytically evaluate the quality of the unsupervised clustering algorithms.

\begin{figure}
    \includegraphics[width=1\linewidth]{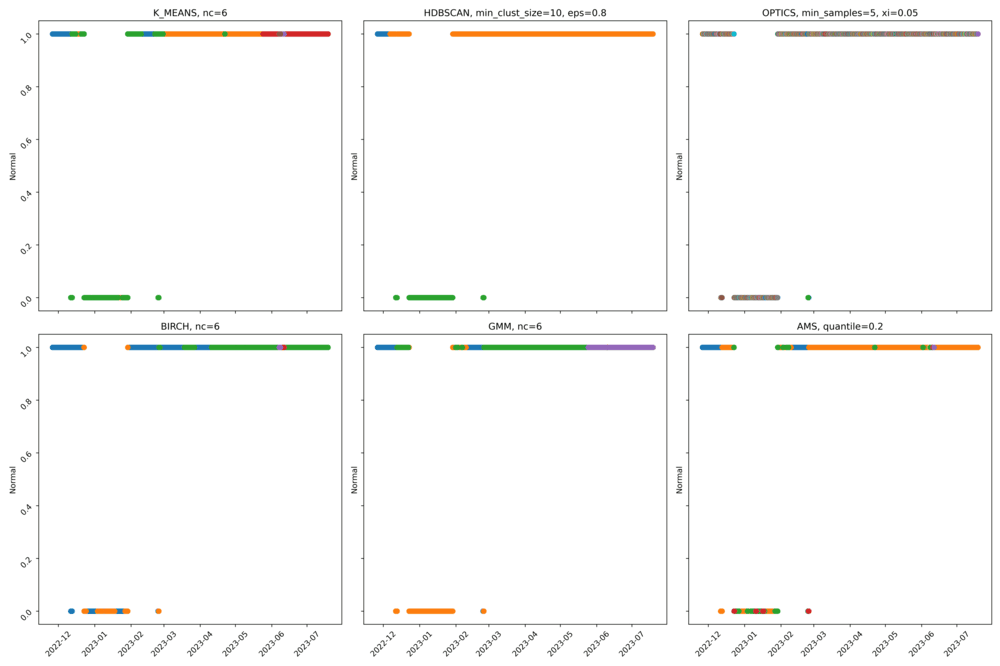}
    \caption{Multiple clustering: The graph shows how different algorithms separate clusters over time, with each point representing an observation colored by its cluster, illustrating the performance and characteristics of each method}
    \label{fig:enter-label}
\end{figure}

\begin{figure}
    \centering
    \includegraphics[width=1\linewidth]{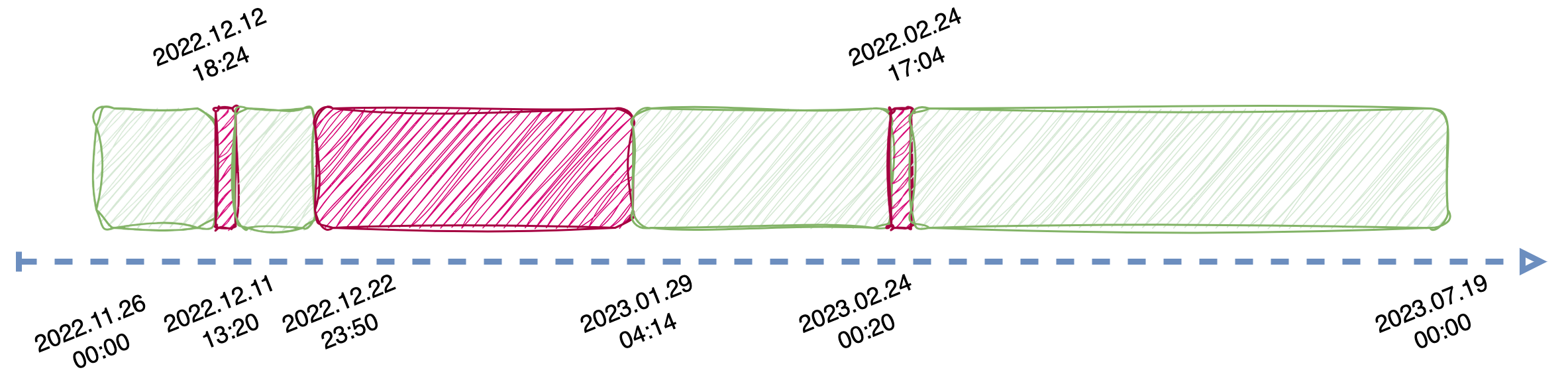}
    \caption{Periods from dataset. Each period has been identified starting from the NoC file}
    \label{fig:enter-label}
\end{figure}

The comparison between Figure 6 and Figure 7 shows that all clustering algorithms identified the three periods of machine abnormality, although the resulting clusters were not entirely pure. Specifically, while the overall patterns of machine abnormality were captured, algorithms like OPTICS and MS-AMS exhibited less distinct cluster boundaries, resulting in mixed clusters. This observation aligns with the metrics for OPTICS, which indicate less clear separation between clusters.

We evaluated the clustering quality by labeling the dataset with known periods from the NoC. To assess the quality of the clustering, we used the Adjusted Rand Index (ARI) and Normalized Mutual Information (NMI) metrics, as outlined in the Strategy description section. ARI measures the agreement between the clustering results and the true labels while accounting for chance, providing insight into the overall cluster coherence. NMI assesses the amount of information shared between the clustering results and the true labels, showing how well the clusters align with the known periods. These metrics, presented in Table 1, were chosen for their ability to capture different aspects of clustering quality and provide a comprehensive assessment of how well each algorithm represents the dataset’s distribution.

\begin{table}[htbp]
    \centering
    \vspace{-10pt} 
    \begin{tabular}{>{\raggedright\arraybackslash}p{0.4\linewidth}p{0.2\linewidth}p{0.2\linewidth}}
        \toprule
        \textbf{CA}& \textbf{ARI} & \textbf{NMI} \\
        \midrule
        \textbf{KMEANS} & 0.30 & \textbf{0.42\textcolor{red}{*}} \\
        \textbf{HDBSCAN} & \textbf{0.55\textcolor{red}{*}} & \textbf{0.64\textcolor{red}{*}} \\
        \textbf{OPTIC} & -0.02 & 0.16 \\
        \textbf{BIRCH} & 0.03 & 0.25 \\
        \textbf{GMM} & \textbf{0.45\textcolor{red}{*}} & \textbf{0.56\textcolor{red}{*}} \\
        \textbf{MS-AMS} & \textbf{0.35\textcolor{red}{*}} & 0.36 \\
        \bottomrule
    \end{tabular}
    \caption{Clustering metrics: ARI (Adjusted Rand Index) and NMI (Normalized Mutual Information). The top three results for each metric are marked with a red asterisk.}
    \label{tab:clustering_metrics}
\end{table}

Because of the analysis performed, we have kept as additional features only those related to the three best metrics (HDBSCAN, GMM, K-MEANS, and MS-AMS), according to the ARI and NMI metrics. The features obtained during the clustering phase are indeed the clusters identified by the specific clustering algorithms considered in our analysis. This decision aims to enhance the ability to identify periods of machine abnormality.

\subsection{Classification models}
Using the NoC information to assign NORMAL/ABNORMAL labels to each observation in the dataset, we proceeded to run the classification models with and without the additional features.
In consideration of the key characteristics of each algorithm, including their learning hypotheses (linear, kernel-based, probabilistic), learning methods (gradient, instance-based, decision tree), and model structure, we identify the following classification algorithms:  (\textcite{davenport2003clustering}):
Logistic Regression 
for linear models,
SVC 
for kernel-based models, GaussianNB 
for probabilistic models, GradientBoostingClassifier 
for gradient-based models, KNeighborsClassifier 
for instance-based models, and RandomForestClassifier 
for decision tree models.

It is important to acknowledge that there may be additional potential subdivisions, and the selection of representatives depends on the specific needs of the addressed problem. Nevertheless, the grouping we have opted for offers a comprehensive overview of the classification algorithms suitable for the dataset of this study.

We leveraged the implementations in sklearn (\textcite{scikitLearn}) for all the algorithms mentioned and executed them with the default parameters.  In our initial modelling step, we implemented the SMOTE technique (Synthetic Minority Over-sampling Technique)  
to effectively tackle class imbalance.

\subsection{Cross-validation and Metrics comparison}
Evaluating the performance of a machine learning model requires assessing its accuracy in both the training set and the test set. The accuracy of the training set illustrates how well the model fits the training data, while the accuracy of the test set provides insights into the model’s performance with new, unseen data. 

We meticulously calculated various metrics for each of the previously mentioned algorithms. Our analysis involved inspecting scenarios with and without pre-clustering and comprehensively considering the execution times in both cases. The ‘Mean Accuracy’ over the train and the test dataset, and the “Mean execution time” during the train process are thoroughly documented in Table 2. Moreover, visual representations of the differences in accuracy and training times between the models are presented in Figure 8 and Figure 9, respectively.

In particular, Figure 8 illustrates how, on average, all clustering models benefit from the use of additional features taken from pre-clustering. Furthermore, the advantage in terms of training time is, occasionally, very significant (Figure 9).

In addition, Table 3 compares the performance of six classification models with and without pre-clustering, focusing on the Recall, F1-score, False Positives (FP), and False Negatives (FN) metrics. 

Logistic Regression (LR) shows similar performance with and without pre-clustering, with a slight improvement in false positives (FP) and false negatives (FN) when pre-clustering is used, indicating enhanced classification accuracy.

The Vector Classifier (SVC) benefits significantly from pre-clustering, achieving a Recall of 0.97 and an F1-score of 0.97, compared to 0.90 and 0.92 without pre-clustering, with increased FP and FN without it.

Gaussian NB (GaussNB) also shows improvement with pre-clustering, with higher Recall and F1-score (0.9673 and 0.96) versus without pre-clustering (0.934 and 0.93), and decreased FP and FN from 7600 and 9100 to 3512 and 3692.

Gradient Boosting (GBM) performs well with a moderate improvement with pre-clustering, showing a Recall and F1-score of 0.9517 and 0.94, compared to 0.925 and 0.92 without pre-clustering.

K-Nearest Neighbors (KNN) significantly improves with pre-clustering, achieving high Recall and F1-score (0.973 and 0.97) and lower FP and FN (2933), compared to a decline without pre-clustering (0.86 and 0.86, with FP and FN at 18,500).

Random Forest (RF) shows excellent performance with pre-clustering, with a Recall of 0.976 and an F1-score of 0.98, improving from 0.92 and 0.92 without pre-clustering, and decreased FP and FN from 10,000 to 2648.

Overall, pre-clustering consistently improves model performance across most classifiers, particularly for K-Nearest Neighbors and Random Forest, showing its effectiveness in enhancing classification accuracy.

\begin{table}[htbp]
    \centering
    \vspace{-10pt} 
    \begin{tabular}{>{\raggedright\arraybackslash}p{0.25\linewidth}p{0.15\linewidth}p{0.15\linewidth}p{0.2\linewidth}}
        \toprule
        \textbf{Algorithm} & \textbf{Mean Accuracy (train)} & \textbf{Mean Accuracy (test)} & \textbf{Mean Execution Time (train)} \\
        \midrule
        \rowcolor{yellow!20} \textbf{SVC} \textcolor{teal}{*}& 1.0000 & 0.9786 & 112.28 s \\
        & 0.9953 & 0.9234 & 732.22 s \\
        \rowcolor{yellow!20} \textbf{LR} \textcolor{teal}{*}& 0.9837 & 0.8037 & 1.28 s \\
        & 0.9019 & 0.7877 & 0.74 s \\
        \rowcolor{yellow!20} \textbf{GaussNB} \textcolor{teal}{*}& 0.9899 & 0.9924 & 0.14 s \\
        & 0.8853 & 0.9344 & 0.11 s \\
        \rowcolor{yellow!20} \textbf{GBM} \textcolor{teal}{*}& 1.0000 & 0.9281 & 96.20 s \\
        & 1.0000 & 0.9240 & 138.25 s \\
        \rowcolor{yellow!20} \textbf{KNN} \textcolor{teal}{*}& 1.0000 & 0.9819 & 1.17 s \\
        & 0.9999 & 0.8585 & 0.86 s \\
        \rowcolor{yellow!20} \textbf{RF} \textcolor{teal}{*}& 1.0000 & 0.9601 & 32.69 s \\
        & 1.0000 & 0.9242 & 57.80 s \\
        \bottomrule
    \end{tabular}   
    \caption{Comparison of algorithms with and without pre-clustering, displaying Mean Accuracy and Mean Execution Time. Rows with a teal asterisk indicate results from pre-clustering features.}
    \label{tab:results1}
\end{table}

\begin{table}[htbp]
    \centering
    \vspace{-10pt} 
    \begin{tabular}{>{\raggedright\arraybackslash}p{0.25\linewidth}llll}
        \toprule
        \textbf{Algorithm} & \textbf{Recall} & \textbf{F1} & \textbf{FP} & \textbf{FN} \\
        \midrule
        \rowcolor{yellow!20} \textbf{SVC} \textcolor{teal}{*} & 0.97 & 0.97 & 2923 & 2923 \\
        & 0.90 & 0.92 & 4500 & 4500 \\
        \rowcolor{yellow!20} \textbf{LR} \textcolor{teal}{*}& 0.97 & 0.97 & 2649 & 2649 \\
        & 0.97 & 0.97 & 15154 & 12765 \\
        \rowcolor{yellow!20} \textbf{GaussNB} \textcolor{teal}{*}& 0.96 & 0.96 & 3512 & 3692 \\
        & 0.93 & 0.93 & 7600 & 9100 \\
        \rowcolor{yellow!20} \textbf{GBM} \textcolor{teal}{*}& 0.95 & 0.94 & 5442 & 5442 \\
        & 0.92 & 0.92 & 8750 & 8750 \\
        \rowcolor{yellow!20} \textbf{KNN} \textcolor{teal}{*}& 0.97 & 0.97 & 2933 & 2933 \\
        & 0.86 & 0.86 & 18500 & 18500 \\
        \rowcolor{yellow!20} \textbf{RF} \textcolor{teal}{*}& 0.97 & 0.98 & 2648 & 2648 \\
        & 0.92 & 0.92 & 10000 & 10000 \\
        \bottomrule
    \end{tabular}   
    \caption{Comparison of algorithms with and without pre-clustering, showing Recall, F1-score, False Positives (FP), and False Negatives (FN). Rows marked with a teal asterisk represent results with pre-clustering features.}
    \label{tab:results2}
\end{table}

\begin{figure}
    \centering
    \vspace{-10pt} 
    \includegraphics[width=1\linewidth]{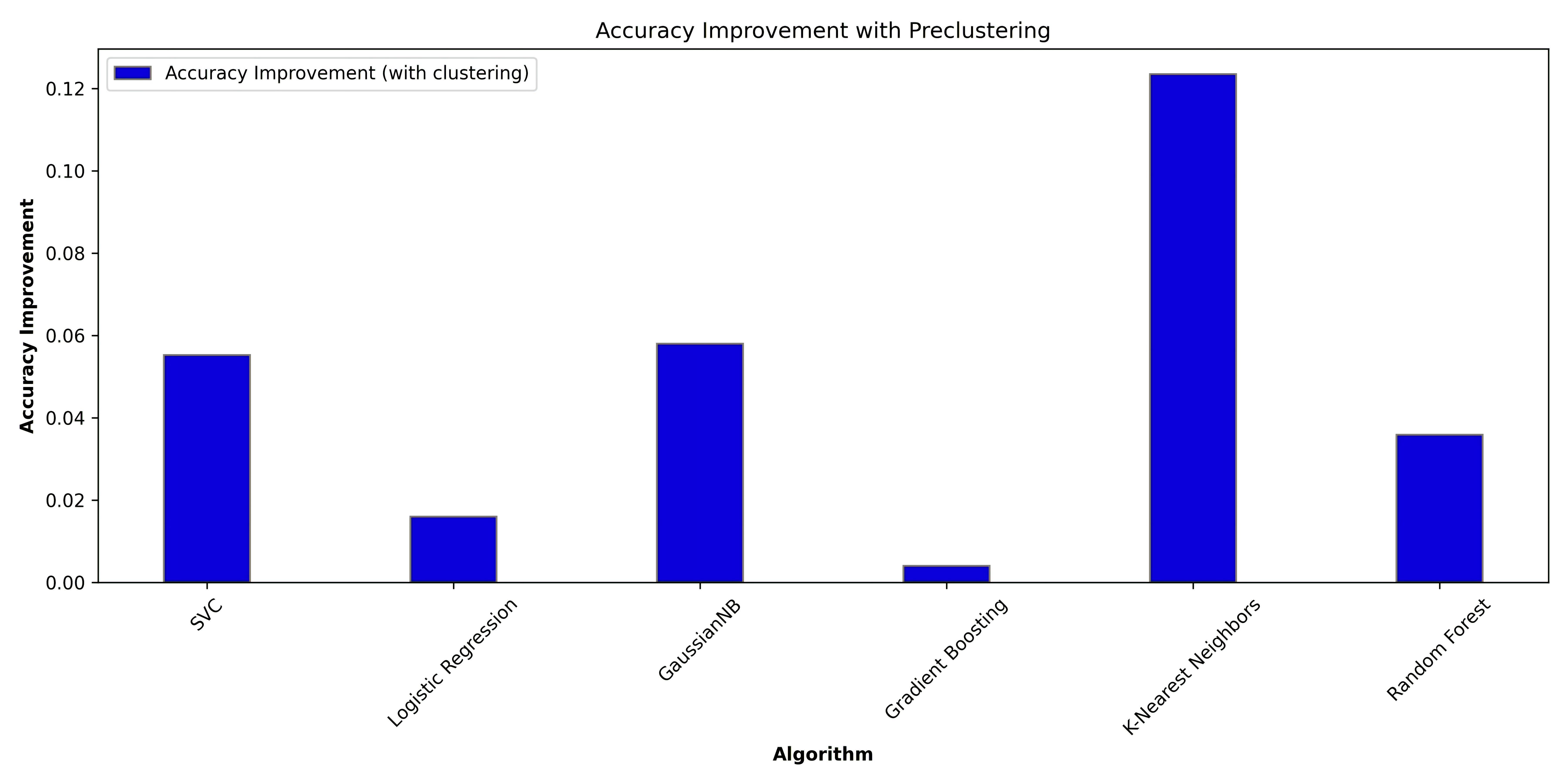}
    \vspace{-10pt} 
    \caption{Accuracy improvement comparison between clustering models with and without pre-clustering features}
    \label{fig:enter-label}
\end{figure}
\begin{figure}
    \centering
    \vspace{-10pt} 
    \includegraphics[width=1\linewidth]{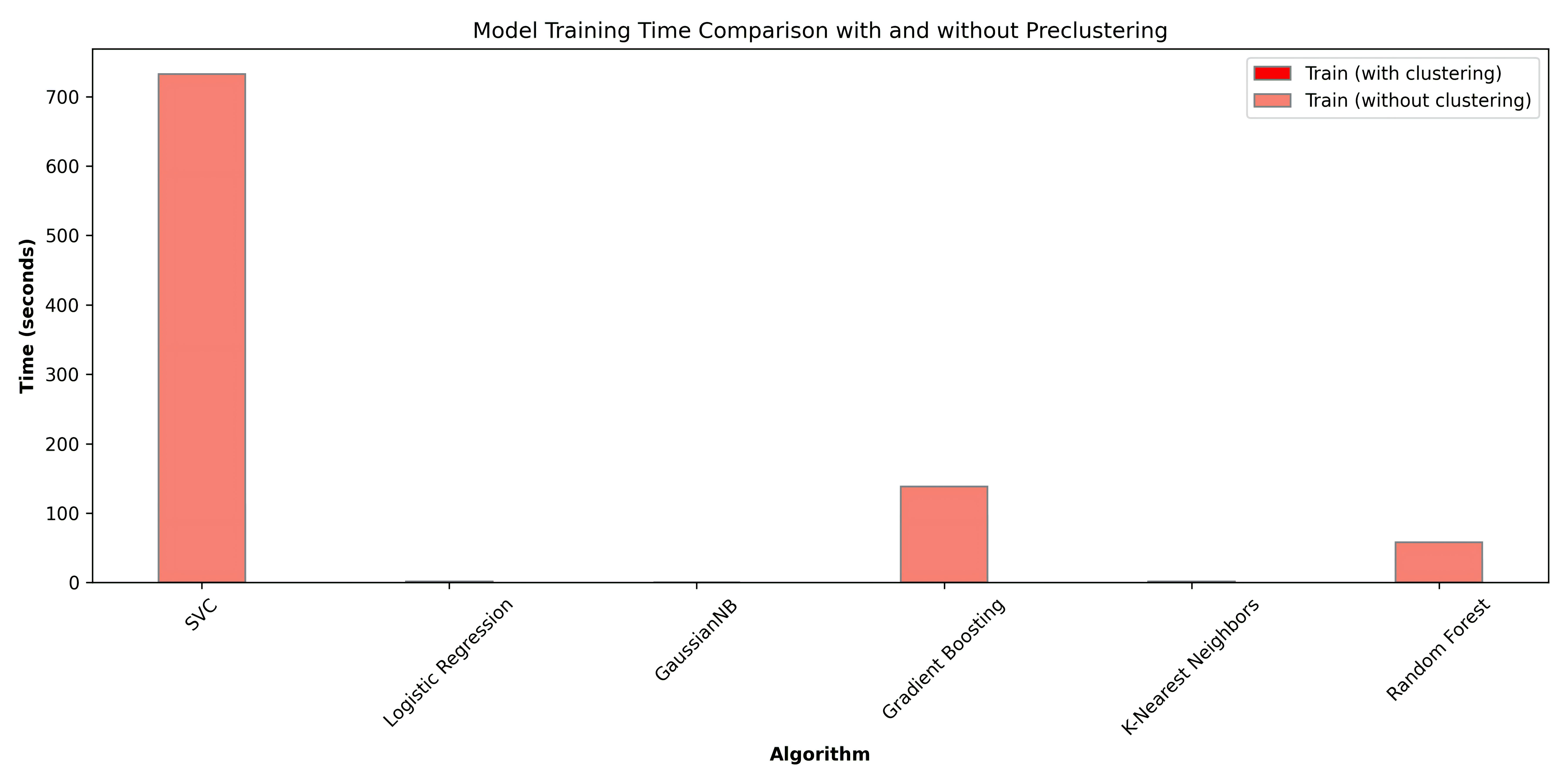}
    \vspace{-10pt} 
    \caption{Comparison of training time reduction with and without pre-clustering features}
    \label{fig:enter-label}
\end{figure}
\section{Results}

The analysis of the results obtained with and without the use of pre-clustering features provides interesting insights into the impact of these features on the accuracy of the clustering models and on training times. About the model accuracy, using pre-clustering features generally improves the accuracy of clustering models on the test set. For example, the Support Vector Classifier (SVC) shows an on-test accuracy of 97.86\% with pre-clustering, compared to 92.34\% without pre-clustering. This pattern of improvement is also observed with other models: the GaussianNB goes from an on-test accuracy of 93.44\% without pre-clustering to 99.24\% with pre-clustering, and the K-Nearest Neighbors (KNN) improves from 85.85\% to 98.19\%. However, not all models follow this trend. Logistic regression, for example, shows only a slight improvement from an on-test accuracy of 78.77\% without pre-clustering to 80.37\% with pre-clustering. In some cases, such as Gradient Boosting and Random Forest, the test accuracy does not change significantly with the use of pre-clustering features, remaining around 92.40\% and 92.42\% respectively. With regards to the training time, the impact of pre-clustering features on training times varies greatly between different algorithms. For the SVC model, the training time with pre-clustering is 112.28 seconds, significantly less than the 732.22 seconds needed without pre-clustering. This suggests that pre-clustering can help reduce computation times for computationally intensive algorithms such as SVC. In contrast, some models show relatively stable training times with and without pre-clustering. For example, logistic regression and GaussianNB show no significant changes in training times. Logistic regression takes approximately 1.28 seconds with pre-clustering and 0.74 seconds without, while GaussianNB shows times of 0.14 seconds with pre-clustering and 0.11 seconds without. In conclusions, using pre-clustering features generally improves the accuracy of the test for most of the models considered. This effect is particularly pronounced for models such as SVC, GaussianNB, and KNN. However, for other models such as Gradient Boosting and Random Forest, the impact on accuracy is less obvious. 
After analyzing the results, it is evident that utilizing the best features identified through unsupervised pre-clustering enhances the accuracy of the models on the test set in identifying potential faults by an average of 4.87\%. The statistical analysis confirmed that this improvement is significant (p-value = 0.0368). Additionally, while the training time for clustering models is reduced by an average of 22.96\%, this reduction did not reach statistical significance (p-value = 0.3104), indicating that the impact on training time may vary depending on the algorithm used.
In the context of a Predictive Maintenance process, these findings highlight the potential of pre-clustering to enhance the accuracy and efficiency of fault detection systems. By improving model accuracy and reducing training times, particularly for computationally intensive algorithms, pre-clustering can contribute to more reliable and timely maintenance interventions, reducing the risk of unexpected machine failures. Compared to the current situation, where predictive models might rely solely on raw data, incorporating pre-clustering can provide a more refined and effective approach to identifying potential faults, making it a valuable addition to Predictive Maintenance strategies.

\section{Discussion}
Some algorithms achieving a training accuracy of 1.0 may indicate overfitting, where models adapt too well to training data but struggle with new data. This could be due to model complexity exceeding the dataset size, leading to memorization rather than generalization. To address this, we could evaluate models on an independent test dataset for a more realistic measure of generalization. Regularization techniques (e.g., L1/L2 regularization, dropout) and restrictive decision tree criteria can help limit model complexity. Expanding the training dataset could also improve model performance. An ablation study could clarify the impact of the pre-clustering phase and parameter variations on effectiveness. Additionally, exploring new feature engineering techniques might enhance data representation and reduce overfitting. Future work should focus on these improvements and re-evaluate model performance on independent data to address overfitting issues.

\subsection{Discussion of Comparisons with Deep Learning Approaches}

Our approach to binary classification should be compared to alternative methods, especially deep learning (DL) models like autoencoders. DL models automate feature extraction and reduce manual preprocessing, but our choice was influenced by the trade-off between accuracy, interpretability, and training time. We plan to include an experiment comparing our method with a DL model using autoencoders to assess accuracy, training time, and model complexity. This comparison will help evaluate the advantages and limitations of different methodologies for binary classification.

\section{Acknowledgement}
Supported by Italian Research Center on High Performance Computing Big Data and Quantum Computing (ICSC), project funded by European Union - NextGenerationEU - and National Recovery and Resilience Plan (NRRP) - Mission 4 Component 2 within the activities of Spoke 2 (Fundamental Research and Space Economy).
We gratefully acknowledge Alfonso Amendola for his technical and scientific consultancy. 
\printbibliography
\end{document}